\pdfoutput=1
\PassOptionsToPackage{prologue,dvipsnames}{xcolor}
\documentclass[11pt]{article}

\usepackage[preprint]{acl}

\usepackage{times}
\usepackage{latexsym}

\usepackage[T1]{fontenc}
\usepackage[utf8]{inputenc}

\usepackage{microtype}

\usepackage{inconsolata}

\usepackage{graphicx}
\usepackage{booktabs}
\usepackage{multirow}
\title{\textsc{EMULATE}: A Multi-Agent Framework for Determining the Veracity of Atomic Claims by Emulating Human Actions}

\author{Spencer Hong \quad Meng Luo \quad Xinyi Wan \\
         National University of Singapore \\
         \texttt{\{spencer.hong, mluo, wan.xinyi\}@u.nus.edu}}

\begin{document}
\maketitle
\begin{abstract}
Determining the veracity of atomic claims is an imperative component of many recently proposed fact-checking systems. Many approaches tackle this problem by first retrieving evidence by querying a search engine and then performing classification by providing the evidence set and atomic claim to a large language model, but this process deviates from what a human would do in order to perform the task. Recent work attempted to address this issue by proposing iterative evidence retrieval, allowing for evidence to be collected several times and only when necessary. Continuing along this line of research, we propose a novel claim verification system, called \textsc{EMULATE}, which is designed to better emulate human actions through the use of a multi-agent framework where each agent performs a small part of the larger task, such as ranking search results according to predefined criteria or evaluating webpage content. Extensive experiments on several benchmarks show clear improvements over prior work, demonstrating the efficacy of our new multi-agent framework. Our code is available at \url{https://github.com/qqqube/EMULATE}.
\end{abstract}

\section{Introduction}

To prevent the spread of misinformation, a multitude of automated fact-checking systems have recently been proposed in the natural language processing community \cite{xie-etal-2025-fire, wang2024factcheck, singal2024evidence, kim2024can, chen2024complex, chern2023factool, pan-etal-2023-fact, wang2023explainable}. For example, \citet{chern2023factool} and \citet{wang2024factcheck} introduced frameworks that break texts into atomic claims and equip LLMs with the ability to use web search tools to retrieve evidence for verifying the claims. Other works have also considered taking iterative approaches for search query generation \citep{wei2024longform} as well as the whole retrieval/verification process \citep{xie-etal-2025-fire}. 

Continuing along this line of research, we propose a novel system, called \textsc{EMULATE}, that takes an atomic claim as input and determines the veracity of the claim by retrieving evidence from the web and mimicking human actions. More specifically, we employ a multi-agent framework that consists of agents for generating search queries, determining the credibility and relevance of search results, evaluating webpage content, assessing collections of evidence, and performing classification. By having each agent execute a small part of the larger task, our system can successfully guide the underlying language models in retrieving important information from external resources which ultimately leads to an amelioration in classification performance. 

The closest work to ours is FIRE \cite{xie-etal-2025-fire}, which consists of three components: one for either outputting the final answer or generating the next search query, another for making web searches and retrieving the snippets of the search results, and a third for final verification after a maximum number of retrieval steps has been reached. Though FIRE also makes use of several agents, our framework further breaks down the fact verification process by trying to understand why additional evidence is needed at each step, which enables the system to ignore redundant and irrelevant search results and generate high-quality queries that can better enhance the system's evidence set.

To evaluate the efficacy of our framework, we perform experiments on a variety of fact-checking benchmarks. Our results show a clear improvement over prior work and demonstrate the effectiveness of using multi-agent systems to tackle complex tasks like fact verification.

\begin{figure*}[t]
  \includegraphics[width=\linewidth]{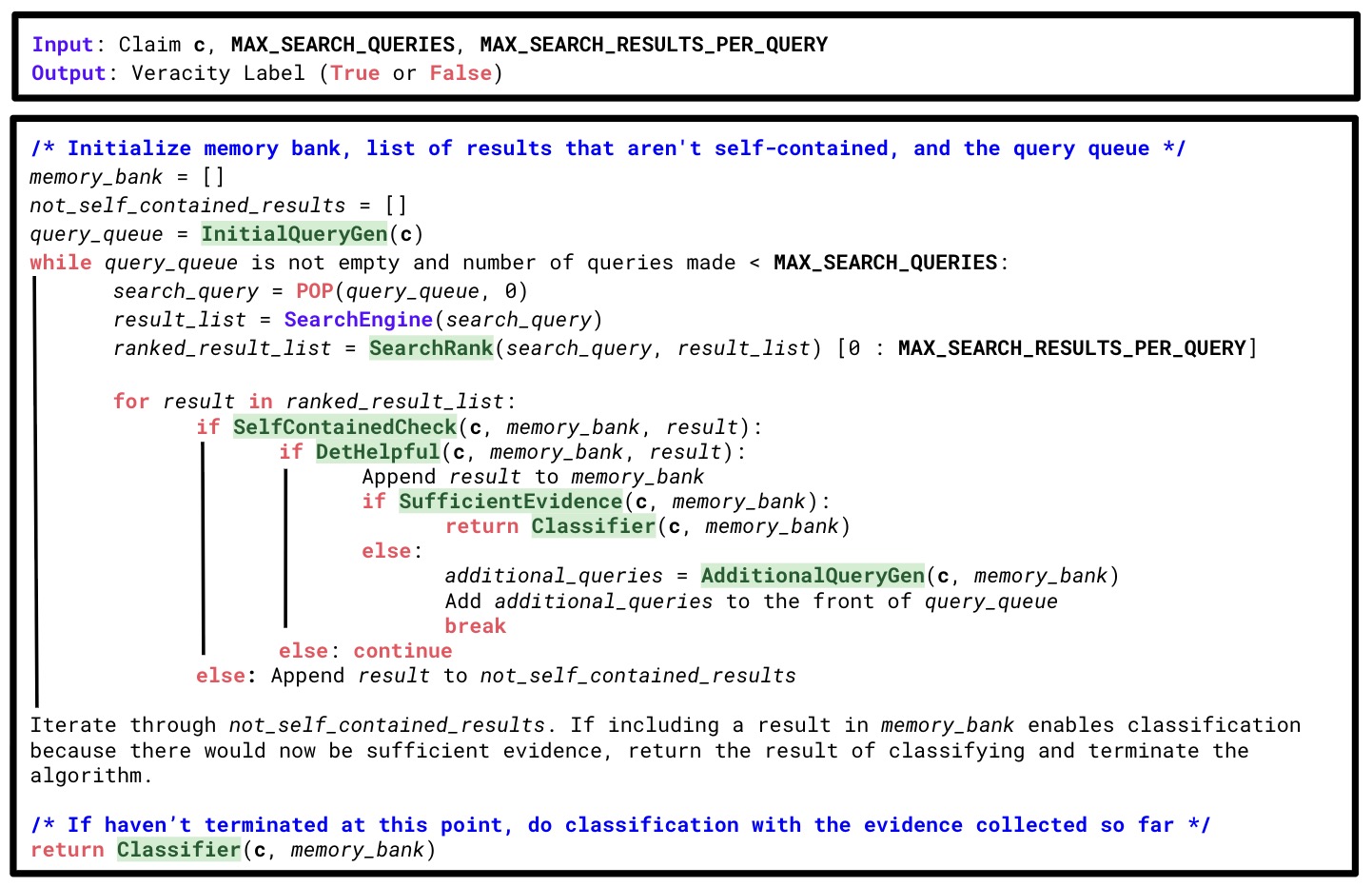}
\vspace{-4mm}
  \caption{\textbf{Claim veracity classification algorithm.} The input to the algorithm is an atomic claim \textbf{c} along with two values specifying the maximum number of search queries that can be made and the maximum number of search results returned per query. The output of the algorithm is a binary label indicating the veracity of the claim. Each LLM agent is highlighted in \colorbox{YellowGreen}{green}. }
  \label{fig:pseudocode}
  \vspace{-4mm}
\end{figure*}

\section{Related Work}

Many existing automatic fact-checking pipelines adopt the \textit{Decompose-Then-Verify} paradigm, which first decomposes a text into several atomic claims and then verifies each claim individually \cite{hu2024decomposition, wei2024longform, song-etal-2024-veriscore, min-etal-2023-factscore}. Several approaches for the latter task of verifying individual claims (which is the task that we focus on in this work), begin by retrieving evidence via web search and then feeding the evidence and the claim to a language model for final verification \cite{chern2023factool, wang2024factcheck}. Though this is sometimes effective, a shortcoming is the misalignment between this process and the process of humans when doing the task. Recent works address this through iterative evidence retrieval \cite{xie-etal-2025-fire}, which allows for evidence to be collected several times and only when it is considered necessary. We build on this idea with \textsc{EMULATE} and also incorporate the idea of iterative retrieval and verification.

\section{Methodology}

\textbf{Emulating Human Actions.} Our multi-agent framework is designed to emulate human actions. If a human were trying to verify a claim by using the Internet, they would start by making a search query that they think would be helpful, which will return many results/links. They would then select a link to click on based on the credibility of the source (which can be inferred from the URL), as well as the relevance (which can be guessed by looking at the title and snippet). After clicking on a link and reading the text, one of the following scenarios will occur:

\textbf{(a)} The document is self-contained and the human has sufficient information to determine the claim's veracity.

\textbf{(b)} The document is self-contained and the human was able to acquire knowledge that’s helpful for the task, but more information is needed.

\textbf{(c)} The document is self-contained, but completely irrelevant.

\textbf{(d)} The document is not self-contained.

If scenario \textbf{(a)} occurs, the human is done with the task. If scenario \textbf{(b)} occurs, the human should retain the information acquired from the text and then think of additional search queries required for completing the task. In scenario \textbf{(c)}, the human should visit another link that was returned in the response to the initial search query. In scenario \textbf{(d)}, the human would need to formulate additional search queries to fill in the gaps. To the best of our knowledge, our system is the first fact-checking algorithm to follow this process.

\noindent\textbf{A Novel Multi-Agent Framework.} Our fact-checking algorithm is shown in Figure~\ref{fig:pseudocode}, and makes use of the following LLM-powered agents:
(1) \textbf{InitialQueryGen:} Generates a list of initial search queries given a claim.
(2) \textbf{SearchRank:} Given a query and a list of corresponding search results (each result consists of a title, a URL, and a snippet), outputs a sorted list of the results based on relevance and credibility. (3) \textbf{SelfContainedCheck:} Given a claim, the evidence set so far, and a new search result, determines if the content of the new webpage is comprehensible (i.e., if it is comprehensible, it is either self-contained or can be comprehended if you consider the information in the evidence set). (4) \textbf{DetHelpful:} Given a claim, the evidence set so far, and a new comprehensible search result, determines if the search result provides new information that isn't already mentioned in the current evidence set and if it would be helpful for veracity checking. (5) \textbf{SufficientEvidence:} Given a claim and the evidence set so far, determines if there is sufficient evidence to perform classification. (6) \textbf{Classifier:} Given a claim and the evidence set, outputs a classification label. (7) \textbf{AdditionalQueryGen:} Given a claim and the evidence set, outputs a list of search queries to enhance the existing evidence set.

\begin{figure}[t]
  \includegraphics[width=0.85\columnwidth]{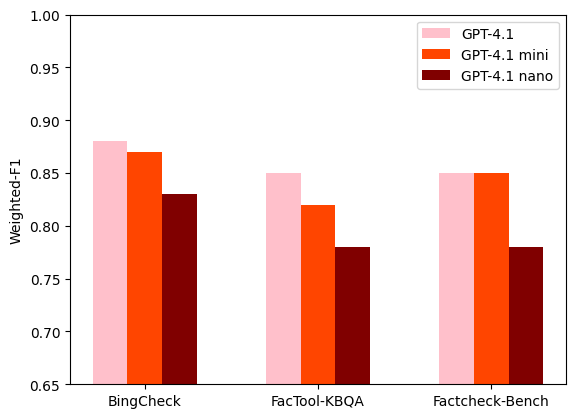}
  \vspace{-3mm}
  \caption{Results on the entire GPT-4.1 model family. The strongest/weakest model according to OpenAI is GPT-4.1/GPT-4.1-nano.}
  \label{fig:whole-family}
  \vspace{-3mm}
\end{figure}

\begin{table}
\centering
\begin{tabular}{l|c|c|c}
\toprule
\textbf{Dataset} & \textbf{\#True} & \textbf{\#False}  & \textbf{Total} \\
\midrule
FacTool-KBQA & 177 & 56  & 233 \\
BingCheck & 160 & 42 & 202 \\
Factcheck-Bench & 472 & 159 & 631 \\
\bottomrule
\end{tabular}
\caption{Dataset statistics for FactTool-KBQA, BingCheck, and Factcheck-Bench}
\label{tab-dataset-stats}
  \vspace{-4mm}
\end{table}

Note that in Figure~\ref{fig:pseudocode}, when the algorithm encounters scenario \textbf{(d)}, it stores the result instead of making additional search queries to fill in the gaps, and walks through them at the end if it didn't terminate during the \textit{while} loop yet, since something that was once not self-contained could become self-contained if the memory bank has changed. This design choice was made to prioritize processing self-contained evidence pieces to minimize the number of queries that need to be made. 

% \begin{figure}[t]
%   \includegraphics[width=0.85\columnwidth]{latex/images/model_scale.png}
%   \vspace{-3mm}
%   \caption{Results on the entire GPT-4.1 model family. The strongest/weakest model according to OpenAI is GPT-4.1/GPT-4.1-nano.}
%   \label{fig:whole-family}
%   \vspace{-4mm}
% \end{figure}

% \begin{table}
% \centering
% \begin{tabular}{l|c|c|c}
% \toprule
% \textbf{Dataset} & \textbf{\#True} & \textbf{\#False}  & \textbf{Total} \\
% \midrule
% FacTool-KBQA & 177 & 56  & 233 \\
% BingCheck & 160 & 42 & 202 \\
% Factcheck-Bench & 472 & 159 & 631 \\
% \bottomrule
% \end{tabular}
% \caption{Dataset statistics for FactTool-KBQA, BingCheck, and Factcheck-Bench}
% \label{tab-dataset-stats}
%   \vspace{-4mm}
% \end{table}

\section{Experiments}

\textbf{Datasets and Metrics.} We evaluate \textsc{EMULATE} along with other systems on three datasets that each provides annotations at the level of atomic claims: FacTool-KBQA \citep{chern2023factool}, BingCheck \citep{li2024self}, and Factcheck-Bench \citep{wang2024factcheck}. FacTool-KBQA is a subset of the dataset introduced in \citet{chern2023factool} for the knowledge-based QA task with 233 claims labeled as either True or False. BingCheck \citep{li2024self} consists of atomic claims annotated with four possible labels (\textit{supported}, \textit{refuted}, \textit{partially supported}, and \textit{not supported}). We retain \textit{supported} and \textit{refuted} examples only and convert their labels to \textit{True} and \textit{False} respectively. We also only use a portion of the \textit{supported} examples to control the class imbalance. Factcheck-Bench \citep{wang2024factcheck} provides 661 checkworthy claims human-annotated with either \textit{True}, \textit{False}, or \textit{Unknown}. We ignore the \textit{Unknown} examples and sample 631 claims for our experiments. See Table~\ref{tab-dataset-stats} for full dataset statistics. 

To quantify performance, we report the precision, recall, and F1 scores for each class. We also provide the macro-F1 score, which aggregates the label-wise F1 scores by averaging. The weighted-F1 score is also included, which could better account for class imbalance.

\begin{table*}[t]
    \centering
    \small 
    \begin{tabular}{c | c | c | c | c | c | c | c | c | c}
    \toprule
    \multirow{2}{*}{\textbf{Dataset}} & \multirow{2}{*}{\textbf{Method}} & \multicolumn{3}{|c|}{\textcolor{blue}{\textbf{True}}} & \multicolumn{3}{|c|}{\textcolor{teal}{\textbf{False}}} & \multirow{2}{*}{\textbf{M-F1} } & \multirow{2}{*}{\textbf{W-F1} } \\
     & & \textbf{P} & \textbf{R} & \textbf{F1} & \textbf{P} & \textbf{R} & \textbf{F1} &  &  \\
    \midrule
    \multirow{5}{*}{BingCheck} & FacTool & \textbf{0.92} &	0.84 &	0.88 &	0.55 &	\textbf{0.71} &	0.62 &	0.75 &	0.83	 \\
    & FactCheck-GPT & \underline{0.91} &	0.8 &	0.85 &	0.48 &	\underline{0.69} &	0.56 &	0.71 &	0.79	 \\
    & SAFE & 0.88 &	0.72 &	0.79 &	0.37 &	0.62 &	0.46 &	0.62 &	0.72 \\
    & FIRE & \underline{0.91} &	\underline{0.87} &	\underline{0.89} &	\underline{0.58} &	\underline{0.69} &	\underline{0.63} &	\underline{0.76} &	\underline{0.84} \\
    & EMULATE & \underline{0.91} &	\textbf{0.96} &	\textbf{0.93} &	\textbf{0.79} &	0.62 &	\textbf{0.69} &	\textbf{0.81} &	\textbf{0.88} \\
    \midrule
    \multirow{5}{*}{FacTool-KBQA} & FacTool & \textbf{0.91} &	0.84 &	0.87 &	0.59 &	\underline{0.73} &	0.65 &	0.76 &	0.82	\\
    & FactCheck-GPT & \textbf{0.91} &	0.77 &	0.83 &	0.51 &	\textbf{0.77} &	0.61 &	0.72 &	0.78	 \\
    & SAFE & 0.89 &	0.87 &	0.88 &	0.61 &	0.64 &	0.63 &	0.76 &	0.82 \\
    & FIRE & \underline{0.9} &	\underline{0.88} &	\underline{0.89} &	\underline{0.63} &	0.68 &	\underline{0.66} &	\underline{0.78} &	\underline{0.83} \\
    & EMULATE & 0.89 &	\textbf{0.92} &	\textbf{0.91} &	\textbf{0.72} &	0.64 &	\textbf{0.68} &	\textbf{0.8} &	\textbf{0.85} \\
    \midrule
    \multirow{5}{*}{Factcheck-Bench} & FacTool & \underline{0.93} &	0.74 &	0.82 &	0.52 &	\underline{0.82} &	0.64 &	0.73 &	0.77	\\
    & FactCheck-GPT & \textbf{0.94} &	0.74 &	0.83 &	0.53 &	\textbf{0.86} &	0.65 &	0.74 &	0.78	 \\
    & SAFE & 0.92 &	0.78 &	0.84 &	0.55 &	0.79 &	0.65 &	0.74 &	0.79 \\
    & FIRE & \underline{0.93} &	\underline{0.81} &	\underline{0.87} &	\underline{0.59} &	0.81 &	\underline{0.68} &	\underline{0.78} &	\underline{0.82}	\\
    & EMULATE & 0.9 & 	\textbf{0.89} &	\textbf{0.9} &	\textbf{0.7} &	0.72 &	\textbf{0.71} &	\textbf{0.8} &	\textbf{0.85}	\\
    \bottomrule
    \end{tabular}
    \vspace{-2mm}
    \caption{For each claim verification system, we report the label-wise precision, recall, and F1 scores along with the Macro-F1 (\textbf{M-F1}) and Weighted-F1 (\textbf{W-F1}) scores. The best results on each dataset are shown in \textbf{bold}, while the second best results are \underline{underlined}.}
    \label{tab:main-results}
    \vspace{-4mm}
\end{table*}

\noindent\textbf{Baselines.} We compare our multi-agent system with four baselines: (1) \textsc{FacTool} \citep{chern2023factool}, (2) \textsc{FactCheck-GPT} \citep{wang2024factcheck}, (3) \textsc{SAFE} \citep{wei2024longform}, and (4) \textsc{FIRE} \citep{xie-etal-2025-fire}. Note that \textsc{FIRE} \citep{xie-etal-2025-fire} is the only baseline that was designed to take as input an atomic claim and output \textit{True} or \textit{False} (like \textsc{EMULATE}). In each of the other three baselines, checking the veracity of atomic claims is one step in the algorithm, which means that minor modifications to the corresponding open-source repositories were required to make comparisons\footnote{For \textsc{FactCheck-GPT}, we also modify the code to utilize \textit{serper.dev} to obtain a maximum of 10 URLs per query.}.

\noindent\textbf{Implementation.} For our main experiments, we employ OpenAI's GPT-4.1 model\footnote{gpt-4.1-2025-04-14} with a temperature of 1 for all agents in \textsc{EMULATE} as well as the baseline systems. All \textsc{EMULATE} agents are provided with zero-shot prompts that contain instructions for the subtasks. Unless otherwise stated, MAX\_SEARCH\_QUERIES and MAX\_SEARCH\_RESULTS\_PER\_QUERY are set to 4 and 2 respectively. To make web searches, we invoke API calls with \textit{serper.dev}.

\begin{table}
  \centering
  \resizebox{\linewidth}{!}{
  \begin{tabular}{ | c | c | c | c | c | c |}
    \hline
    \textbf{Dataset} & \textbf{Method} & \textbf{\textcolor{blue}{True} F1} & \textbf{\textcolor{teal}{False} F1} &  \textbf{Weighted-F1} \\
    \hline
    \multirow{3}{*}{FacTool-KBQA} & RM-SR & 0.87 & 0.57 & 0.8 \\
    & RM-SCC & 0.9 & 0.63 & 0.84 \\
    & \textsc{EMULATE} & 0.91 & 0.68 & 0.85 \\
    \hline
     \multirow{3}{*}{Factcheck-Bench} & RM-SR & 0.88 & 0.68 &	0.83	\\
    & RM-SCC & 0.88 & 0.66 &	0.82 \\
    & \textsc{EMULATE} & 0.9 &	0.71 & 0.85 \\
    \hline
  \end{tabular}}
  \vspace{-2mm}
  \caption{Ablation studies on FacTool-KBQA and FactCheck-Bench. RM-SR/RM-SCC means that \textbf{SearchRank}/\textbf{SelfContainedCheck} were removed from \textsc{EMULATE}.}
  \label{tab:ablations}
\end{table}
\vspace{-2mm}

\section{Results}

Our main results are presented in Table~\ref{tab:main-results}. From them, we can see that \textsc{EMULATE} outperforms all baselines on every dataset on 6 out of 8 metrics that we compute. Notably, \textbf{\textsc{EMULATE} consistently achieves the best results on both label-wise F1 scores, the macro-F1 score, and the weighted-F1 score}, which confirms the effectiveness of our novel design. We also observe that \textsc{FIRE} always achieves the second best results, which is likely attributed to its iterative retrieval mechanism.

To gain a better understanding of the impact that different agents have on our system, we conduct ablation studies on FacTool-KBQA and FactCheck-Bench. In particular, we quantify the effect of removing (1) \textbf{SearchRank} and (2) \textbf{SelfContainedCheck}. From Table~\ref{tab:ablations}, we can see that excluding \textbf{SearchRank} leads to performance degradation on both datasets (more heavily on FacTool-KBQA), which tells us that the \textbf{SearchRank} agent can effectively sort a list of search results according to the aforementioned criteria. We also find degradation on all datasets when removing the \textbf{SelfContainedCheck} agent, which reveals that the agent can effectively evaluate and filter search results.

Lastly, we run experiments on the entire GPT-4.1 model family to determine if \textsc{EMULATE} still works well when the underlying LLM of each agent is supplanted with a weaker model. According to Figure~\ref{fig:whole-family}, as the underlying LLM weakens, the weighted-F1 scores decrease as well. Intuitively, weaker models are expected to be less performant on the subtasks in \textsc{EMULATE}, which can lead to suboptimal results; however, we can see that the performance of \textsc{EMULATE} when equipped with GPT-4.1-mini is sometimes close to the performance with GPT-4.1.

\section{Conclusion}

In this paper, we proposed a novel approach for determining the veracity of atomic claims, which is designed to emulate human actions through a multi-agent framework. Through extensive experiments, we showed that our system, \textsc{EMULATE}, outperforms previously introduced algorithms for the task and can work well even when used with a weaker base LLM. We also reported the results from doing ablation studies, which confirmed the effectiveness of several agents. 

\section*{Limitations}

Evaluation of our system requires datasets that have veracity annotations at the level of atomic claims. Due to the scarcity of such datasets, we were only able to evaluate on three, and each contained less than 1,000 examples. Additionally, these datasets have a class imbalance issue (i.e., there are significantly less \textit{False} claims than \textit{True} claims).

Another shortcoming lies in our design choice of processing documents that aren't self-contained at the end of the algorithm. Future work should investigate other alternatives, since for some claims, it may not be possible to do claim verification without providing search results that aren't self-contained as evidence.

% Bibliography entries for the entire Anthology, followed by custom entries
%\bibliography{anthology,custom}
% Custom bibliography entries only
\bibliography{custom}

\begin{thebibliography}{13}
\providecommand{\natexlab}[1]{#1}

\bibitem[{Chen et~al.(2024)Chen, Kim, Sriram, Durrett, and Choi}]{chen2024complex}
Jifan Chen, Grace Kim, Aniruddh Sriram, Greg Durrett, and Eunsol Choi. 2024.
\newblock Complex claim verification with evidence retrieved in the wild.
\newblock In \emph{Proceedings of the 2024 Conference of the North American Chapter of the Association for Computational Linguistics: Human Language Technologies (Volume 1: Long Papers)}, pages 3569--3587.

\bibitem[{Chern et~al.(2023)Chern, Chern, Chen, Yuan, Feng, Zhou, He, Neubig, Liu et~al.}]{chern2023factool}
I~Chern, Steffi Chern, Shiqi Chen, Weizhe Yuan, Kehua Feng, Chunting Zhou, Junxian He, Graham Neubig, Pengfei Liu, and 1 others. 2023.
\newblock Factool: Factuality detection in generative ai--a tool augmented framework for multi-task and multi-domain scenarios.
\newblock \emph{arXiv preprint arXiv:2307.13528}.

\bibitem[{Hu et~al.(2024)Hu, Long, and Wang}]{hu2024decomposition}
Qisheng Hu, Quanyu Long, and Wenya Wang. 2024.
\newblock Decomposition dilemmas: Does claim decomposition boost or burden fact-checking performance?
\newblock \emph{arXiv preprint arXiv:2411.02400}.

\bibitem[{Kim et~al.(2024)Kim, Lee, Huang, Chan, Li, and Ji}]{kim2024can}
Kyungha Kim, Sangyun Lee, Kung-Hsiang Huang, Hou~Pong Chan, Manling Li, and Heng Ji. 2024.
\newblock Can llms produce faithful explanations for fact-checking? towards faithful explainable fact-checking via multi-agent debate.
\newblock \emph{arXiv preprint arXiv:2402.07401}.

\bibitem[{Li et~al.(2024)Li, Peng, Galley, Gao, and Zhang}]{li2024self}
Miaoran Li, Baolin Peng, Michel Galley, Jianfeng Gao, and Zhu Zhang. 2024.
\newblock Self-checker: Plug-and-play modules for fact-checking with large language models.
\newblock In \emph{Findings of the Association for Computational Linguistics: NAACL 2024}, pages 163--181.

\bibitem[{Min et~al.(2023)Min, Krishna, Lyu, Lewis, Yih, Koh, Iyyer, Zettlemoyer, and Hajishirzi}]{min-etal-2023-factscore}
Sewon Min, Kalpesh Krishna, Xinxi Lyu, Mike Lewis, Wen-tau Yih, Pang Koh, Mohit Iyyer, Luke Zettlemoyer, and Hannaneh Hajishirzi. 2023.
\newblock \href {https://doi.org/10.18653/v1/2023.emnlp-main.741} {{FA}ct{S}core: Fine-grained atomic evaluation of factual precision in long form text generation}.
\newblock In \emph{Proceedings of the 2023 Conference on Empirical Methods in Natural Language Processing}, pages 12076--12100, Singapore. Association for Computational Linguistics.

\bibitem[{Pan et~al.(2023)Pan, Wu, Lu, Luu, Wang, Kan, and Nakov}]{pan-etal-2023-fact}
Liangming Pan, Xiaobao Wu, Xinyuan Lu, Anh~Tuan Luu, William~Yang Wang, Min-Yen Kan, and Preslav Nakov. 2023.
\newblock \href {https://doi.org/10.18653/v1/2023.acl-long.386} {Fact-checking complex claims with program-guided reasoning}.
\newblock In \emph{Proceedings of the 61st Annual Meeting of the Association for Computational Linguistics (Volume 1: Long Papers)}, pages 6981--7004, Toronto, Canada. Association for Computational Linguistics.

\bibitem[{Singal et~al.(2024)Singal, Patwa, Patwa, Chadha, and Das}]{singal2024evidence}
Ronit Singal, Pransh Patwa, Parth Patwa, Aman Chadha, and Amitava Das. 2024.
\newblock Evidence-backed fact checking using rag and few-shot in-context learning with llms.
\newblock In \emph{Proceedings of the Seventh Fact Extraction and VERification Workshop (FEVER)}, pages 91--98.

\bibitem[{Song et~al.(2024)Song, Kim, and Iyyer}]{song-etal-2024-veriscore}
Yixiao Song, Yekyung Kim, and Mohit Iyyer. 2024.
\newblock \href {https://doi.org/10.18653/v1/2024.findings-emnlp.552} {{V}eri{S}core: Evaluating the factuality of verifiable claims in long-form text generation}.
\newblock In \emph{Findings of the Association for Computational Linguistics: EMNLP 2024}, pages 9447--9474, Miami, Florida, USA. Association for Computational Linguistics.

\bibitem[{Wang and Shu(2023)}]{wang2023explainable}
Haoran Wang and Kai Shu. 2023.
\newblock Explainable claim verification via knowledge-grounded reasoning with large language models.
\newblock In \emph{Findings of the Association for Computational Linguistics: EMNLP 2023}, pages 6288--6304.

\bibitem[{Wang et~al.(2024)Wang, Reddy, Mujahid, Arora, Rubashevskii, Geng, Afzal, Pan, Borenstein, Pillai et~al.}]{wang2024factcheck}
Yuxia Wang, Revanth~Gangi Reddy, Zain Mujahid, Arnav Arora, Aleksandr Rubashevskii, Jiahui Geng, Osama~Mohammed Afzal, Liangming Pan, Nadav Borenstein, Aditya Pillai, and 1 others. 2024.
\newblock Factcheck-bench: Fine-grained evaluation benchmark for automatic fact-checkers.
\newblock In \emph{Findings of the Association for Computational Linguistics: EMNLP 2024}, pages 14199--14230.

\bibitem[{Wei et~al.(2024)Wei, Yang, Song, Lu, Hu, Huang, Tran, Peng, Liu, Huang, Du, and Le}]{wei2024longform}
Jerry Wei, Chengrun Yang, Xinying Song, Yifeng Lu, Nathan~Zixia Hu, Jie Huang, Dustin Tran, Daiyi Peng, Ruibo Liu, Da~Huang, Cosmo Du, and Quoc~V Le. 2024.
\newblock \href {https://openreview.net/forum?id=4M9f8VMt2C} {Long-form factuality in large language models}.
\newblock In \emph{The Thirty-eighth Annual Conference on Neural Information Processing Systems}.

\bibitem[{Xie et~al.(2025)Xie, Xing, Wang, Geng, Iqbal, Sahnan, Gurevych, and Nakov}]{xie-etal-2025-fire}
Zhuohan Xie, Rui Xing, Yuxia Wang, Jiahui Geng, Hasan Iqbal, Dhruv Sahnan, Iryna Gurevych, and Preslav Nakov. 2025.
\newblock \href {https://aclanthology.org/2025.findings-naacl.158/} {{FIRE}: Fact-checking with iterative retrieval and verification}.
\newblock In \emph{Findings of the Association for Computational Linguistics: NAACL 2025}, pages 2901--2914, Albuquerque, New Mexico. Association for Computational Linguistics.

\end{thebibliography}

\appendix

%\section{Example Appendix}
%\label{sec:appendix}

%This is an appendix.

\end{document}